# Dense-Cast: A lightweight ensemble of deep learning architectures for precipitation nowcasting

**Gourav Jyoti Kalita[1], Hidam Kumarjit Singh[1]**

[1]Department of Electronics and Communication Technology, Gauhati University, Guwahati-781014, Assam, India

**Abstract:** Proper short-term forecasting of precipitation is crucial in disaster management and preparedness. Nonetheless, the variability and nonlinearity of precipitation make short-term forecasting challenging for meteorologists. Moreover, capturing temporal dependencies in spatiotemporal data is a challenge in precipitation nowcasting. In this article, we introduce a lightweight deep learning model for half-hourly precipitation nowcasting. This model has been designed by incorporating the DenseNet architecture, residual connections, and transformer encoders for effective precipitation nowcasting with reduced model parameters. The North-Eastern region of India has been selected as the area of interest for our study. The region receives the highest precipitation during the months of June-September due to the monsoon season. The proposed model takes the previous five time-steps of half-hourly precipitation as inputs and predicts the precipitation in the next two half-hours. The GPM IMERG precipitation dataset with a 30-minute cadence has been used in this study for training and testing the model. The proposed architecture achieves best MAE of 0.235 millimetres, RMSE of 0.735 millimetres, and KGE score of 0.816 at an interval of 30 minutes.

## 1. Introduction

Precipitation is a vital element for regulating the Earth's climate and sustaining life. It is the primary pathway for water returning to the atmosphere, influencing freshwater availability, agricultural productivity and ecological balance. Additionally, precipitation influences the global water and energy cycles, impacting atmospheric circulation and soil moisture. Precipitation may vary from gentle rainfall to heavy rainfall. Light and medium rainfall nourish Earth's flora and fauna while heavy rainfall leads to natural hazards, such as flash floods, landslides, and urban waterlogging. Effective short-term precipitation forecasting is crucial for disaster management and preparedness. In the field of meteorology, nowcasting pertains to the prediction of a weather variable with a lead time of up to 6 hours. With the evolution of artificial intelligence (AI) technology and the abundance of spatiotemporal meteorological data, deep learning has achieved state-of-the-art performance in forecasting complex weather processes[1]. However, with the widespread use of AI technology, building lightweight solutions nowadays aligns with the sustainable development goals, which pertains to a significant reduction in power consumption, computational resources and time [2, 3].

Numerical weather prediction (NWP) models are commonly used for weather forecasting purposes. However, these models have several drawbacks, including such as high computational costs, longer computation time and low noise immunity [3]. In addition to NWP models, optical flow algorithms, which track the motion of pixels across consecutive frames, are also utilised for nowcasting tasks. However, optical flow algorithms have several disadvantages, such as the inability to account for intensity changes, dependency on image sequences, and limitations in modelling dynamic changes [4, 5]. Data-driven deep learning (DL) algorithms have recently gained popularity among meteorologists for short-term forecasting due to their capability to extract nonlinear patterns from data, better noise immunity, low cost, and fast execution.

In this article, we are proposing a lightweight deep learning architecture that combines dense blocks, residual architectures, and a transformer encoder for half-hourly precipitation nowcasting. Here, the model leverages the capabilities of depth-wise separable convolutions, enforced in dense and residual blocks, which enable the construction of lightweight architectures while preserving the memory efficiency and representational mapping obtained from the data.

Based on the nature of the forecast, weather forecasting is divided into two types - ***deterministic*** and ***probabilistic***. Deterministic forecasting focuses on predicting continuous values of the weather variable, whereas probabilistic forecasting focuses on the different probability distributions. Probabilistic forecasting is often preferred in various scenarios due to its ability to incorporate uncertainty and provide a range of possible outcomes. However, it is important to note that this approach has its limitations. Challenges in effective communication of the results can arise, and it tends to be more computationally intensive as compared to deterministic methods. In this context, precipitation nowcasting is viewed as a deterministic image-to-image translation task. Here, the model takes the previous five precipitation images of the study area as input to predict the precipitation for the upcoming two half-hour intervals. Although several meteorological variables that correlate with precipitation are used as predictors, this study focuses solely on historical precipitation data.

## 2. Literature Review

In recent times, many researchers have been exploring lightweight DL architectures tailored for various computer vision applications, particularly in the area of precipitation nowcasting. An overview of pertinent literature that relates to our study is given below.

Howard AG et al. 2017 introduced a new class of models called MobileNets for computer vision applications in mobile phones [6]. The authors introduced separable convolutions, a lightweight version of convolutional neural networks that help build efficient deep learning models for computer vision.

Gao Huang et al. 2018 developed the DenseNet architecture which consists of densely connected convolutional neural networks [7]. This method helps in mitigating the vanishing gradient problem in feed-forward neural networks, and enable feature reuse.

Ionescu et al. 2021 proposed a lightweight precipitation nowcasting model for satellite products based on the Xception architecture [8]. This model is a lightweight architecture that takes five satellite products as inputs and produces the precipitation outputs of the next one to three time-steps. The proposed model yields effective results as compared to a baseline CNN model with fewer parameters. However, it was unable to outperform RNN-based models, which is a significant drawback.

Li et al. 2023 proposed an ensemble framework of multiple lightweight architectures to tackle problems encountered in DL based nowcasting methods [9]. This method was reported to be able to tackle problems caused by diverse patterns of precipitation, data unavailability and imbalance arising out of sparsity of weather data.

M Sit et. al. 2024 developed a rainfall nowcasting model based on the MobileNet-V2 architecture [10]. The method was reported to be able to generate forecasts for the next 120 minutes based on precipitation data from the previous 120 minutes. This method is parameter-efficient and simpler as compared to some of the deep learning architectures.

Kalita et al. 2025, [11] introduced a lightweight U-Net based model for deterministic precipitation nowcasting at an hourly interval during the monsoon season over North-East India. This model has nearly four times lesser parameters than the vanilla U-Net model, and gives effective foreasts with best $R^2$ score of 0.718 with the ERA5 reanalysis hourly dataset.

## 3. Materials and Methods

### 3.1 *Study area*

Our research concentrates on rainfall forecasting over the North-East region of India, which includes eight distinct states: Arunachal Pradesh, Assam, Manipur, Meghalaya, Mizoram, Nagaland, Sikkim, and Tripura. This diverse area is situated between longitudes of 87° E and 98° E, and latitudes of 21° N to 30° N. North-East India is a geographically diverse region with the Brahmaputra and Barak being the main river basins in the region. Also, the region lies in the Himalaya and Indo-Burma biodiversity hotspot, where dense rainforests, sprawling wetlands, and steep mountains provide a home to countless species. During June to September, the region receives the maximum precipitation due to the monsoon season [12]. The monsoon arrives in the region as a blessing for local farmers, flooding the agricultural fields and rivers every year. This brings nutrient-rich river sediments, which enhance agricultural productivity [13, 14]. The rain-bearing southwest monsoon, which originates from the Bay of Bengal, brings thunderstorms and heavy rainfall in the area. This results in various natural hazards, including flash floods and river erosion in the plains, particularly in the Brahmaputra River basin, as well as landslides in the mountainous regions. Additionally, unexpected thunderstorms and hail in the region cause damage to crops, livestock and human livelihoods. Moreover, floods in the region also destroy animal shelters in the reserved forests and sanctuaries and cause grave suffering to them as well [15]. Effective short-term precipitation forecasting is crucial for mitigating disaster risks and saving lives in these situations.

We have defined a bounding square that encompasses the region from 87° E to 95° E longitude and 20° N to 33° N latitude. Along with the northeastern states of India, this area encompasses the entirety of Bhutan and Bangladesh, as well as a few portions of Myanmar and Tibet. Figure 1 illustrates a sample of the precipitation map over the selected area chosen for our experiment.

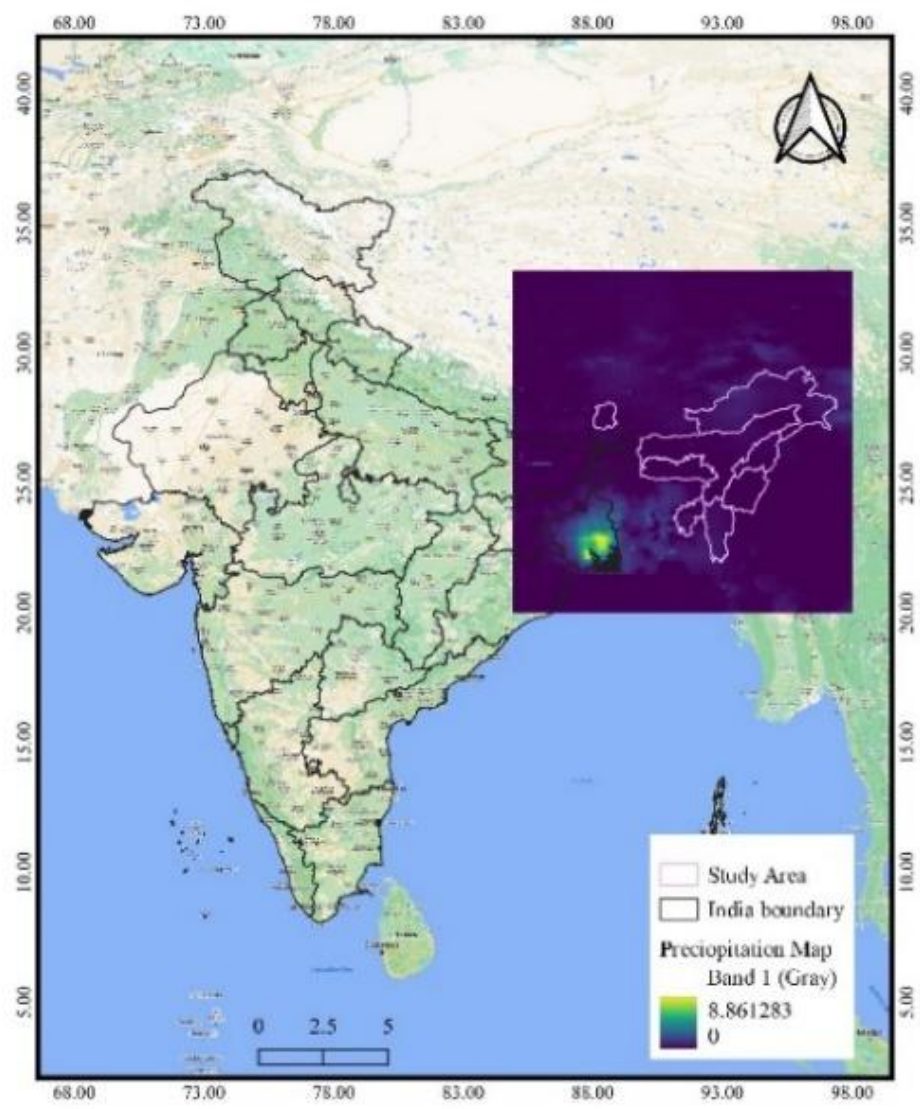

Figure 1: The study area and a sample of the data utilized for this research.

### 3.2 Dataset

The precipitation variable of the IMERG Final Precipitation V07 dataset from NASA's Global Precipitation Measurement (GPM) mission [16] at a half-hourly interval is used for the experiment. The GPM IMERG is a high-resolution dataset with a spatial resolution of $0.1^{o}$ x $0.1^{o}$ at a 30-minute cadence. The dataset combines multi-sensor data, which includes passive microwave satellite instruments, microwave-calibrated infrared satellite instruments, and ground-based precipitation gauges. The data fusion approach in the dataset makes it unique and more accurate, which helps to overcome the limitations of a single sensor. The IMERG dataset is available in three stages of data: ***The Early Run****, the* ***Late Run****, and* ***the Final Run****.* Here, the Early Run provides data approximately four hours after the observation, the Late Run has a latency of 14 hours after observation, while the Final Run is available approximately 3.5 months from the month of observation.

The half-hourly precipitation dataset from June to September for five consecutive years, 2019 to 2023, was used for training and testing the models in our study.

### 3.3 Data preparation and preprocessing

Before fitting the data to a model, the data goes through several preprocessing steps, which include: finding missing timestamps, filling gaps, data cleaning, cropping images, normalization, and train-test split. The preprocessing steps used in the study are presented in Table 1.

Table 1: Summary of the preprocessing steps used in the experiment.

| Preprocessing | Description |
|---|---|
| **Finding and filling missing data** | Search for missing timestamps, and if any missing timestamps are found, recollect the missing data to fill the gaps |
| **Reducing redundancy** | If any duplicate data is found, delete the redundant data |
| **Data cleaning** | NA values are replaced by the mean value or zero. Also, if there are any negative values, they are replaced by zero since precipitation cannot have negative values. |
| **Cropping images** | Since the data are fitted to our model as images, the images are cropped to a size of 128 x 128 pixels for proper fitting. |
| **Normalization** | The data are normalised to a range of 0 and 1, using the min-max scaler. |
| **Creating features and labels** | The precipitation maps of five consecutive half hours are stacked across the channels of the images. These are used as features. While the next two half-hours of precipitation are stacked as features. |
| **Train-test split** | 80 percent of the data is used as the training data, while the rest is used as the testing dataset. |

### *3.4 Proposed lightweight DL model*

The proposed is an ensemble of multiple architectures. This model features an encoder-decoder architecture that includes a temporal transformer bridge block. The model harnesses the abilities of separable convolutions employed in dense blocks, residual blocks and transformers for effective precipitation nowcasting. The building blocks of this model are described below.

***(i)*** ***Encoder:*** We employ Residual blocks and dense blocks to build the encoder of our model, enabling it to extract features from the input data effectively. Depth-wise separable convolutions [6] are used as the basic layer of these blocks. Figure 2 illustrates the structural diagrams of the residual architecture, the basic dense layer, the dense block, and the transition block used in the encoder.

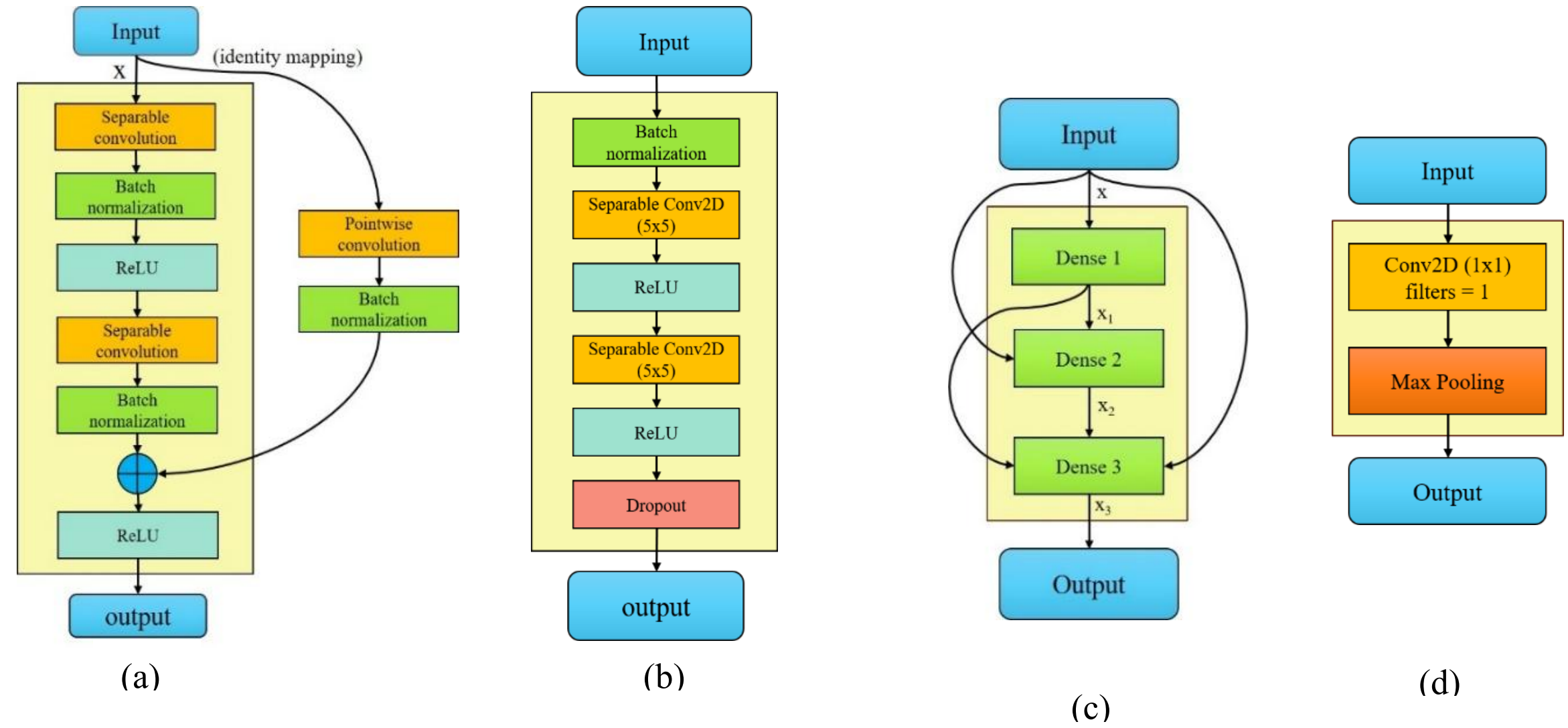

Figure 2: The structural diagrams of the blocks used in the encoder. 2(a) the residual block, 2(b) the basic dense layer, 2(c) the dense block and 2(d) the transition layer.

***(ii)*** ***Residual block:*** The lightweight residual blocks used here are taken from [11]. The separable convolutions, batch normalization, and the rectified linear unit (ReLU) activation in the residual block are used to extract the features directly from the data, while the shortcut path with the pointwise convolutions and batch normalization adds the original features with the learned features, which helps to alleviate the vanishing gradient problem occurring in large convolutional feedforward architectures. The mappings obtained by the residual blocks are represented by equation (1).

$$Y_n = F(X_n, \{W_i\}) + W_s X_n \tag{1}$$

Here, $X_n$ and $Y_n$ represents the inputs and the outputs of the n$^{th}$ block, respectively, while, $W_i$ and $W_s$ represents the weights. The first part of equation 1 represents the mapping obtained by the feedforward part of the block, while the second part represents the weights and the identity mapping through the shortcut path. These are finally added to produce the overall residual mapping.

***(iii)*** ***Dense block:*** The dense block used here is a modified lightweight version of the original dense blocks proposed by the authors of the original DenseNets paper [7]. In contrast to residual architectures, DenseNets address the vanishing gradient problem found in deep convolutional networks by concatenating features learned from each layer with those of the next one. This dense connectivity guarantees efficient gradient flow and feature reuse, as each layer can access both newly learned representations and all previously learned feature maps through concatenation. This feature of the DenseNets makes it lighter compared to the ResNets. The mapping obtained by the dense blocks is represented by equation (2).

$$Y_n = H_j([x_0, x_1, \dots, x_{j-1}]) \tag{2}$$

Here, $Y_n$ represents the overall mapping obtained by the dense block, while $x_0, x_1, \ldots, x_{j-1}$ represents the mapping obtained by each layer inside the dense block. $H_j(.)$ represents the nonlinear transformation applied to each layer, while [.] represents the concatenation of the features extracted by each dense layer. Here,$i$ represents the depth of the dense block equal to the number of dense layers within the dense block, and here we are utilising a depth of, $j = 3$.

*(iv)* ***Transition block:*** The transition block used here is composed of a pointwise convolution layer followed by a 2x2 max pooling layer, which reduces the size of the images to half. This helps the model to generalise better and avoid overfitting by reducing the feature size. This block reduces dimensionality, controls complexity and maintains efficient gradient flow between blocks.

*(v)* ***Bridge:*** A ***transformer encoder*** is used as ***the bridge*** in the architecture to capture the temporal relationship between the meteorological data. As compared to the existing temporal networks such as the LSTM, GRU, and ConvLSTM, the transformer has some advantages, such as the ability to capture long-term temporal dependencies, better gradient flow and less bias towards recent events, making it more efficient and reliable for time series forecasting[17,18].

The transformer architecture utilises the multi-head attention mechanism to capture the temporal dependencies from the data. The transformer takes the features learned from the last encoding block as input, and the data is reshaped into some tokens, which are a sequence of patches of the images. The transformer learns from the tokens and attempts to understand which regions and timestamps are most influential, which helps us achieve a better forecast. Figure 3 shows the structural diagram of the transformer encoder architecture.

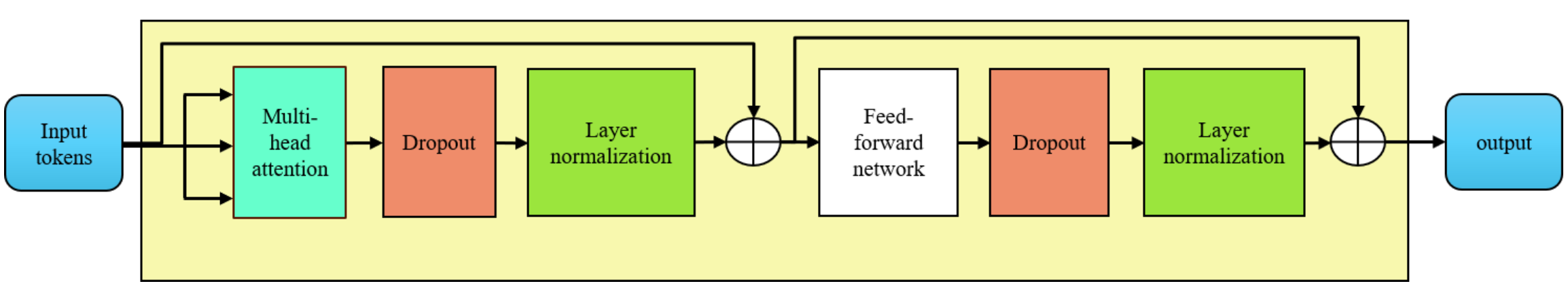

Figure 3: The structural diagram of the transformer encoder used as the bridge in the model for temporal mapping.

The mappings obtained by the transformer are represented by equations (3) and (4).

$$\text{Transformer output, } T(X) = (X_{ffn} + A_T) \tag{3}$$

$$\text{Where, } A_T = (X + Z(X)) \tag{4}$$

Here, $T(X)$ represents the mapping obtained by the transformer encoder, $A_T$ represents the output of the first part of the encoder, which includes the multi-head attention output and the residual connection. $Z(X)$ represents the mapping obtained by the multi-head attention while, $X_{ffn}$ represents the mapping obtained by the feed-forward network with the layer normalization and the dropout applied additionally.

***(vi)*** ***Decoder:*** It comprises transpose convolutions and residual blocks that reconstruct the precipitation maps based on the features learned from the encoded data. The reconstructed features by the decoder are concatenated with the corresponding encoder block for better localisation of features. The block diagram of the decoder is represented in Figure 4(b). The function of the decoder is expressed as in equation (5):

$$X_{dec} = (Residual(Concat(X^T, f_{enc})) \tag{5}$$

Here, the mapping reconstructed by the decoder is represented by $X_{dec}$ while $X^T$ represented the features obtained by the transposed convolution operation and $f_{enc}$ represents the corresponding encoded features.

The block diagrams of the encoder block, the decoder block and the bridge block are illustrated in Figure 4.

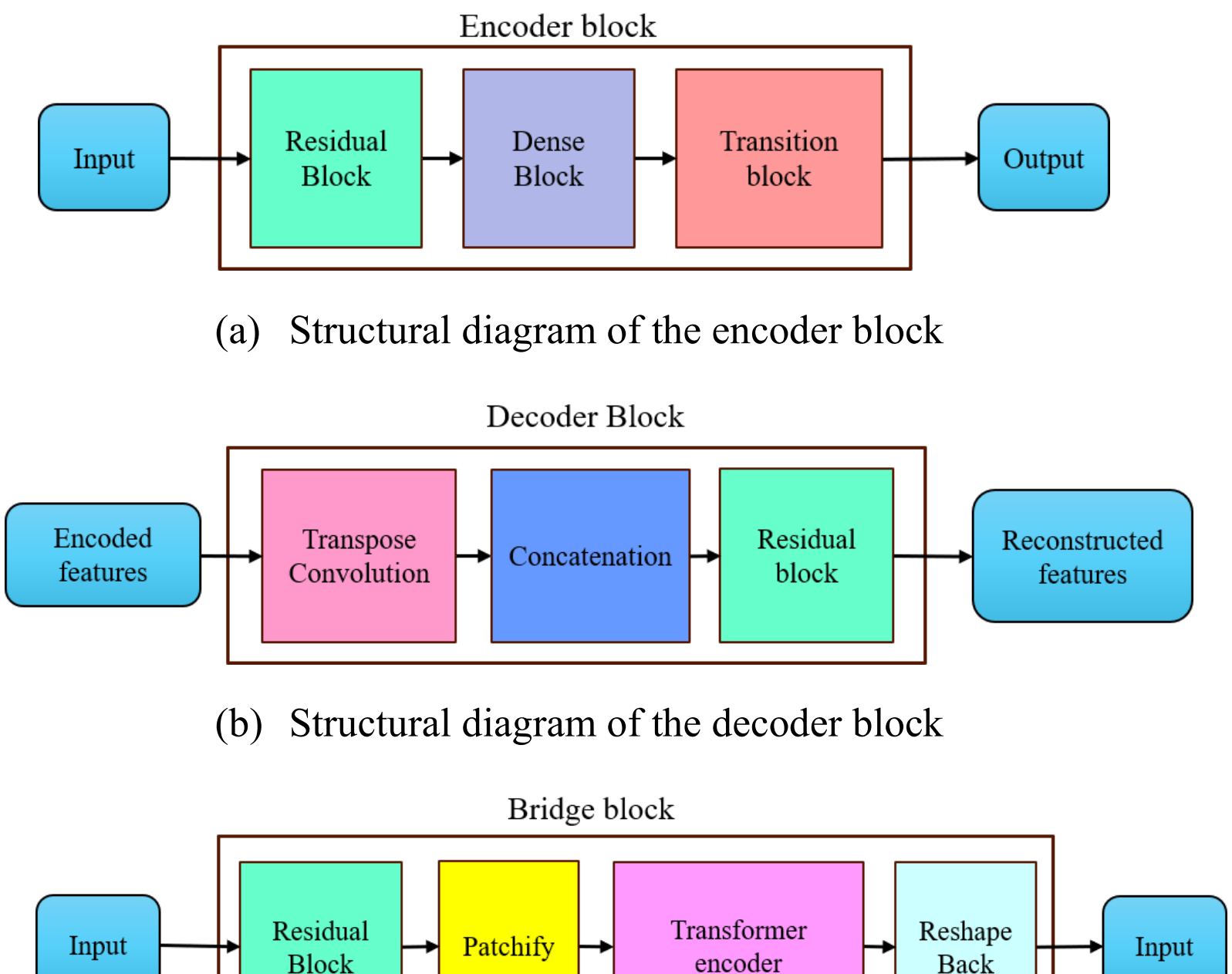

Figure 4: Structural diagrams of the (a) the encoder, (b) the decoder and (c) the bridge block used in the model.

***(vii)*** ***Output layer*:** At the final stage of the model, a pointwise convolution layer with a sigmoid activation function is applied to produce the desired output, with the number of filters matching the required number of time steps. It can be expressed as in equation (6):

$$Y = \sigma(Conv2D_{(2\times2)}(X_{dec})) \quad (6)$$

Here, $Y$ represents the predicted output by the model, and $\sigma$ represents the sigmoid activation function applied at the output layer, which generalizes the values within a range of 0 and 1.

Figure 5 illustrates the structural diagram of the proposed model architecture. The proposed model is a deep architecture with a total of six encoding layers, a bridge block, and seven decoding layers. Here, the model takes five consecutive frames of half-hourly precipitation, i.e., t to t-4, and produces the precipitation maps for t+1 and t+2.

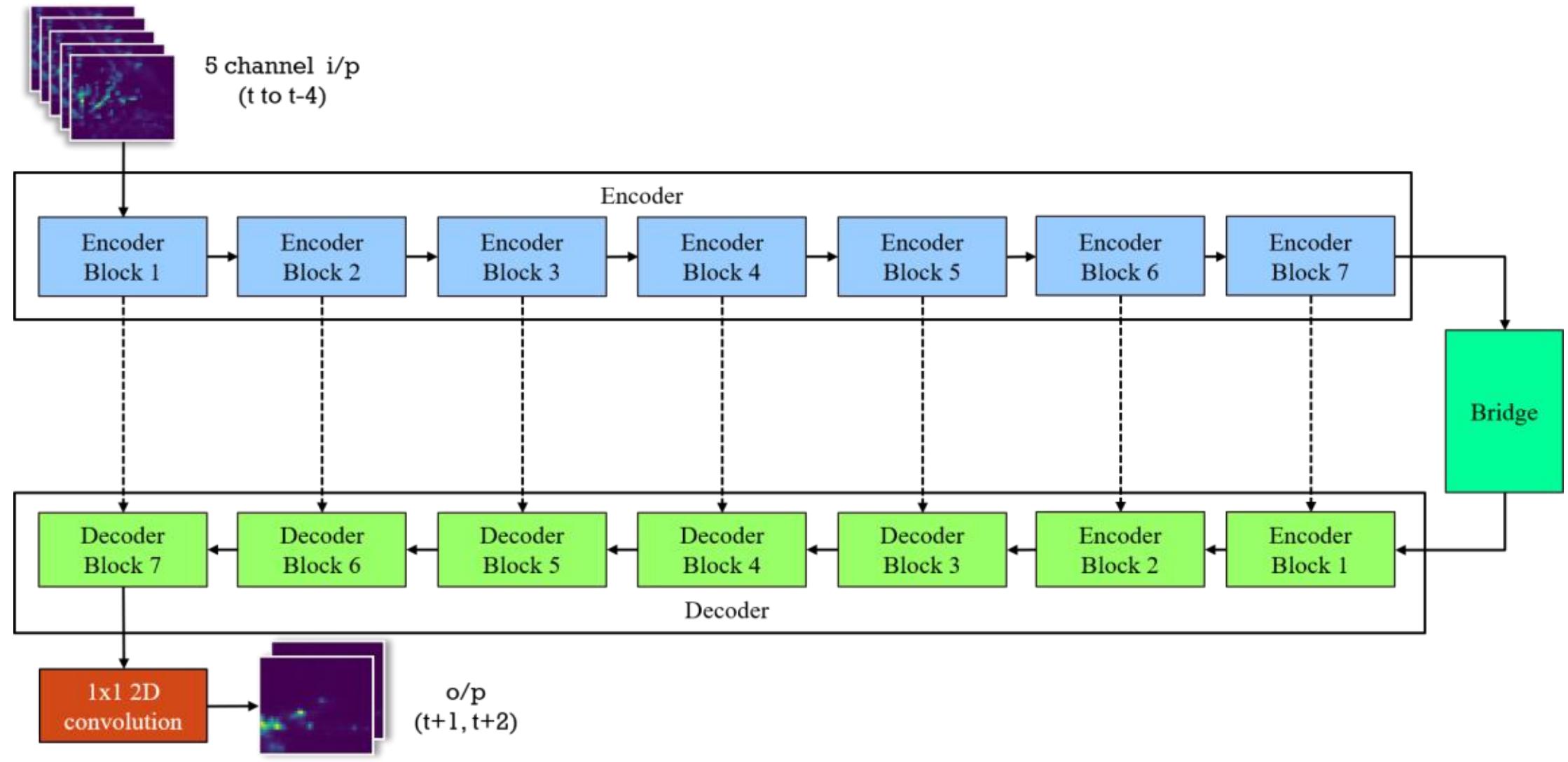

Figure 5: Structural diagram of the proposed model architecture.

### 3.5 Experiment details

Each encoder block extracts the features as channels equal to the number of filters applied, as shown in Figure 5. Here, a multiplier is used equal to the depth of the model. A growth rate of 16 is applied to the model, which is sufficient for a DenseNet based model to function properly [7]. In the transformer block, 32 filters are applied, resulting in optimal performance. The half precipitation dataset for the months of June-September of the years 2018 to 2023 is used as the dataset. The size of the dataset equals to a total of 29,280 images. 80 percent of the dataset is used for training and validation of the model, while the remaining 20 percent is used for testing. No augmentation is applied in the dataset during the training and testing process. The model is trained with an initial learning rate of 0.0001 with the Adam optimizer, the Mean Squared Error loss function, and a batch size of 16. A learning rate decay is applied, which reduces the learning rate after five consecutive epochs by a factor of 0.2. If no progress in the loss is observed, the training is stopped after 30 epochs. The training and testing of the model is a trial-and-error process, which is repeated until an optimal model is obtained. The training of the model has been executed for 100 epochs. The model is trained and tested in a Python environment in Google Colab Pro with the Nvidia A100 GPU runtime.

Mean Absolute Error (MAE), Root Mean Squared Error, Kling Gupta Efficiency (KGE) are used for evaluating the performance of our model. Here, the MAE and the RMSE are used to evaluate the quantitative efficiency of the model, while KGE is used for evaluating the hydrological significance of the model. Table 2 shows a summary of the evaluation metrics used in the experiment.

Table 2: Summary of the statistical metrics used for evaluation of the model predictions.

| Statistical Metric | Mathematical expression | Definition |
|---|---|---|
| **MAE** | $MAE = \frac{1}{n} \sum \lvert Y_{actual} - Y_{predicted} \rvert$ | The mean of the absolute difference between the actual value and the predicted value. It ranges from 0 to ∞. The lesser the MAE, the better the model predictions. |
| **RMSE** | $RMSE = \sqrt{\frac{\sum_{i=1}^{N} (Y_{actual} - Y_{predicted})^2}{n}}$ | The square root of the mean of the squared difference between the actual values and the predicted values. It ranges from 0 to ∞. The lesser its value, the better the model predictions. |
| **KGE** | $KGE = 1 - \sqrt{(r-1)^2 + (\alpha - 1)^2 + (\beta - 1)^2}$ Here, r is the Pearson correlation coefficient, while $\alpha$ is the ratio of the standard deviation of simulated and observed values, which represents the variability of prediction errors. $\beta$ is a bias term which equals the ratio of the mean of simulated to observed values. | It was developed by Gupta et al. 2009 [19] to improve the existing metrics such as the MSE and the Nash-Sutcliffe Efficiency for hydrological modelling. It ranges from $-\infty$ to 1. The value of KGE closer to 1 represents a better model performance. |

## 4. Results and Discussions

The propose DL model has been evaluated by using the statistical metrics described in Table 2, and the performance of the model is compared with a U-Net model with similar depth and configuration. Evaluated performance metrics of the model has been shown in Table 3. From Table 3, it can be seen that the proposed model produces a best KGE score of 0.816 at a lead time of 30 minutes. With a lesser parameters equal to half of the U-Net model, the proposed model produces a better KGE score. Although the MSE and RMSE produced by the model are slightly higher than those of the U-Net model, the higher KGE score signifies that the model is able to capture the variability and correlation effectively. The proposed model, with significantly fewer model parameters, produces competing results with the U-Net, which is a significant merit of the present work.

Table: 3. Summary of the evaluation results of the model.

| Lead time | Model | Total parameters | MAE (mm) | RMSE (mm) | KGE |
|---|---|---|---|---|---|
| 30 min | Dense-Cast | 3.3 M | 0.235 | 0.735 | **0.816** |
| | U-Net | 7 M | **0.195** | **0.677** | 0.774 |
| 1 hour | Dense-Cast | 3.3 M | 0.262 | 0.837 | **0.708** |
| | U-Net | 7 M | **0.238** | **0.776** | 0.692 |

The higher KGE score indicates that the model is suitable for capturing hydrological dynamics; however, the model still requires improvements in terms of overall accuracy, which can be achieved by fine-tuning it more precisely. Figure 6 shows some samples of the precipitation maps predicted by our model against the ground truth data.

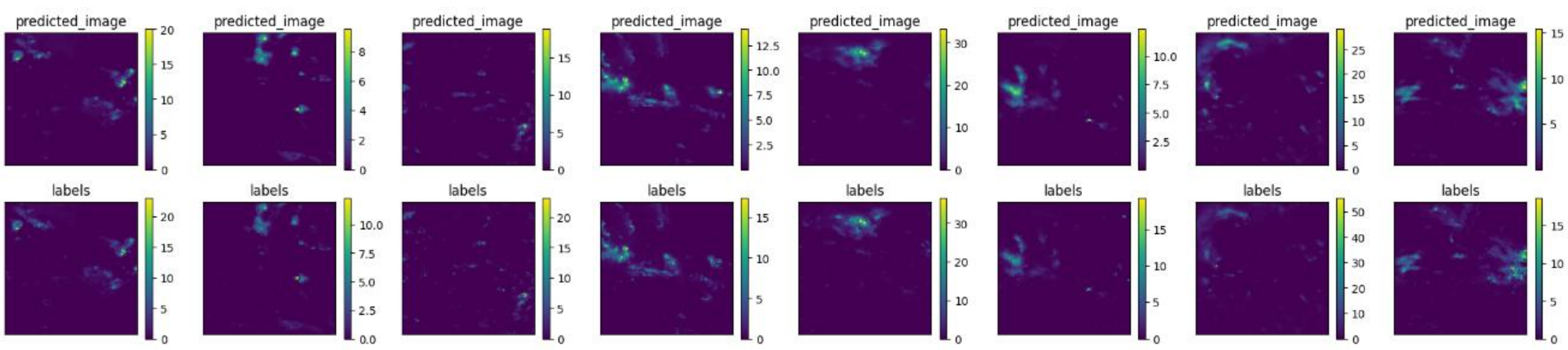

Figure 6: Samples of the precipitation maps anticipated by the proposed model

## 5. Conclusion and future prospects

In this article, we have explored how to build computationally cost-effective DL models for precipitation nowcasting using multiple lightweight architectures. Here, we employed the transformer-based attention mechanism to leverage the efficiency of a CNN-based encoder-decoder model. Evaluation of the model's efficiency by testing with two different lead times, as evident from the KGE score, shows that the model is effective for both the 30-minute and one-hour intervals. During the hours of natural hazards or extreme weather events, the forecast from a reliable forecasting model can be a crucial input. Lightweight and cost-effective DL architectures such as Dense-Cast can be a very helpful tool in such situations.

However, the model uses only historical precipitation data as inputs; in future, we aim to use more variables as inputs that correlate with causes of precipitation in the area. Moreover, the present study employs a reanalysis dataset only. Future direction of this work will employ real-time satellite and radar datasets for real-time nowcasting.